\documentclass[11pt]{article}

\usepackage[final]{acl}

\usepackage{times}
\usepackage{latexsym}

\usepackage[T1]{fontenc}

\usepackage[utf8]{inputenc}

\usepackage{microtype}

\usepackage{inconsolata}

\usepackage{graphicx}
\usepackage{amsmath,amssymb,amsfonts}
\usepackage{makecell}
\usepackage{multirow}
\usepackage{pifont}
\usepackage{arydshln}
\usepackage{tcolorbox}
\tcbuselibrary{breakable}

\definecolor{darkgreen}{rgb}{0.0, 0.5, 0.0}
\definecolor{darkred}{RGB}{164.0, 11.0, 7.0}

\title{SpiralThinker: Latent Reasoning through an Iterative Process with Text–Latent Interleaving}

\author{Shengmin Piao \\
  Yonsei University \\
  Seoul, South Korea \\
  \texttt{shengminp@yonsei.ac.kr} \\
  \And
  Sanghyun Park\textsuperscript{\textdagger} \\
  Yonsei University \\
  Seoul, South Korea \\
  \texttt{sanghyun@yonsei.ac.kr} \\}

\begin{document}
\maketitle
\renewcommand{\thefootnote}{}
\footnotetext{\textsuperscript{\textdagger} Corresponding author.}
\renewcommand{\thefootnote}{\arabic{footnote}}
\begin{abstract}
Recent advances in large reasoning models have been driven by reinforcement learning and test-time scaling, accompanied by growing interest in latent rather than purely textual reasoning. However, existing latent reasoning methods lack mechanisms to ensure stable reasoning dynamics in latent space and a systematic way to interleave implicit and explicit reasoning. 
We introduce SpiralThinker, a stabilized iterative latent reasoning framework that performs iterative updates over latent representations while interleaving latent and textual reasoning steps. At its core, it combines a progressive alignment objective that explicitly regulates latent representations across iterations with structured annotations for text–latent interleaving, thereby stabilizing latent updates and maintaining coherence with textual reasoning.
Across mathematical, logical, and commonsense reasoning tasks, SpiralThinker achieves state-of-the-art performance among latent reasoning baselines. Further analysis shows that both iteration and alignment are essential, that the optimal numbers of latent tokens and iterations vary by dataset, and that proper alignment is crucial for effective iterative latent reasoning. Overall, SpiralThinker bridges iterative computation and latent reasoning, demonstrating that aligned iterative updates can reliably steer reasoning in the latent space. Code is available at \href{https://github.com/shengminp/SpiralThinker}{https://github.com/shengminp/SpiralThinker}.
\end{abstract}

\section{Introduction}
Recent advances in large reasoning models have been driven primarily by progress in reinforcement learning \cite{guo2025deepseek} and test-time scaling \cite{snell2024scaling}, which have improved models' reasoning capability to generate longer and more intricate chains of thought \cite{wei2022chain}. In parallel, another line of research explores latent reasoning, where reasoning unfolds within high-dimensional hidden representations rather than explicit textual sequences \cite{hao2024training}. Despite their apparent differences, both paradigms share the same objective: enriching a model’s internal computation, either explicitly through text tokens or implicitly through hidden representations\footnote{Throughout this paper, latent (or implicit) reasoning refers to reasoning that occurs within hidden representations, whereas textual (or explicit) reasoning denotes reasoning expressed through textual tokens.}.

Existing latent reasoning methods show promise in modeling the underlying reasoning trajectory \cite{chen2025reasoning, zhu2025survey, li2025implicit} but remain underexplored in two critical respects.

\begin{figure}[t]
  \centering
  \includegraphics[width=1.0\columnwidth]{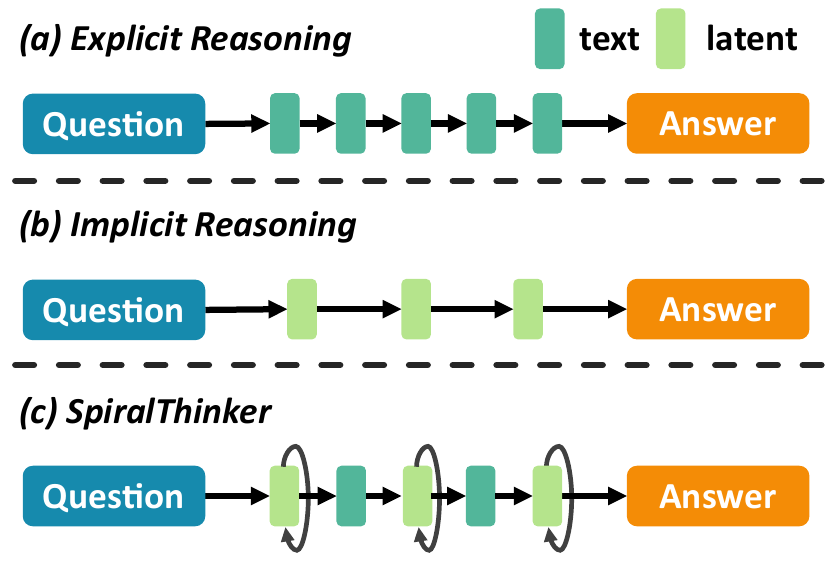}
  \caption{\textbf{(a)} Explicit reasoning processes textual tokens once. \textbf{(b)} Implicit reasoning processes latent representations once. \textbf{(c)} SpiralThinker interleaves textual with latent reasoning via an iterative process.}
  \label{fig:overall framework}
\end{figure}

One underexplored aspect concerns the mechanisms for performing stable and coherent reasoning in latent space. Most approaches treat latent representations as token-level inputs \cite{goyal2024think, hao2024training, shen2025codi} and process them in a single forward pass \cite{zhu2025survey}. However, this token-centric perspective forces latent representations to encode entire reasoning steps simultaneously, redistributing computation without modeling how reasoning unfolds in latent space. We argue that latent reasoning should be explicitly modeled as an iterative process, wherein latent representations are progressively enhanced to support subsequent reasoning and the final answer.

Another unexplored aspect involves how to interleave explicit and implicit reasoning at the step level. Human reasoning naturally alternates between internal thinking and external expression, whereas textual-only frameworks verbalize every step, risking either overthinking \cite{chen2025do} or underthinking \cite{wang2025thoughts}. Latent approaches, in contrast, compress the reasoning steps into latent representations \cite{deng2023implicit, hao2024training, shen2025codi}, thereby sacrificing interpretability and controllability. Bridging these paradigms requires a method capable of transitioning between interpretable textual reasoning and compact latent reasoning.

To address these gaps, we propose \textit{SpiralThinker} (Figure \ref{fig:overall framework}), a stabilized iterative latent reasoning framework that performs implicit reasoning through iterative refinement of latent representations while systematically integrating it with explicit reasoning.
At each iteration, SpiralThinker updates latent representations to deepen the reasoning trajectory without generating additional tokens, enabling extended implicit reasoning within a standard autoregressive decoding process.

To ensure coherent and goal-directed latent reasoning, we introduce a progressive alignment objective that constrains latent representations to remain consistent with their explicit textual counterparts throughout the iterative process. This objective stabilizes iterative updates and promotes coherent information flow throughout the iterative process. 
Furthermore, SpiralThinker employs a structured annotation scheme that formalizes text–latent interleaving as an alternating reasoning process, guiding the model on when to alternate between explicit and implicit reasoning steps.
Together, these mechanisms yield deeper yet more coherent latent reasoning trajectories, effectively bridging explicit interpretability with implicit computational depth.

We evaluate SpiralThinker on three datasets: GSM8K-Aug \citep{deng2023implicit}, ProsQA \citep{hao2024training}, and StrategyQA \citep{geva-etal-2021-aristotle}, which cover mathematical, logical, and commonsense reasoning tasks. Experimental results show that SpiralThinker consistently outperforms existing latent reasoning methods and adapts robustly to diverse reasoning tasks.

Empirically, ablation studies on the iterative process and the alignment objective highlight their complementary roles in enabling stable iterative reasoning. We further analyze the impact of the number of latent tokens, iterations, and the weighting schedule, revealing dataset-specific optima for the first two and confirming that appropriate alignment is essential for effective latent reasoning with the iterative process.

The key contributions of this work are as follows:
\begin{itemize}
    \item We introduce SpiralThinker, a stabilized iterative latent reasoning framework that explicitly regulates latent updates across iterative refinement steps and formalizes structured text–latent interleaving within a unified reasoning process.
    \item We design a progressive alignment objective that provides iteration-aware supervision for latent representations across refinement steps, maintaining coherence with the corresponding textual reasoning and ensuring stable, goal-directed latent updates.
    \item We identify dataset-specific optima for the number of latent tokens and iterations, and demonstrate that proper alignment is crucial for achieving effective latent reasoning with the iterative process.
\end{itemize}

\section{Related Work}
\subsection{Latent Reasoning}
In the current paradigm dominated by decoder-only architectures, latent reasoning primarily manifests in the model’s internal representations—namely token embeddings, hidden states, and output logits—serving as the substrates where reasoning unfolds implicitly.

Early attempts introduced a latent token represented by a learnable embedding that is excluded from the language modeling objective. This design preserves the semantics of the original vocabulary while delaying output generation to extend internal computation \cite{goyal2024think, herel2024thinking, pfau2024lets}. Subsequent works treated the embeddings associated with reasoning steps as continuous information carriers, applying compression techniques to obtain compact representations that can be integrated into the language modeling objective \cite{cheng2024compressed, su2025token, xu-etal-2025-softcot}. 

Shifting focus from token embeddings to the model’s internal hidden states, researchers explored how to provide effective learning signals. Some studies employed curriculum learning to gradually internalize reasoning capabilities \cite{deng2024explicit, liu2024can}, while others aligned hidden states of latent tokens with corresponding explicit reasoning steps, thereby enabling the model to process latent tokens as internal analogues of textual reasoning \cite{deng2023implicit, shen2025codi}.

\begin{figure*}[t]
  \centering
  \includegraphics[width=0.9\linewidth]{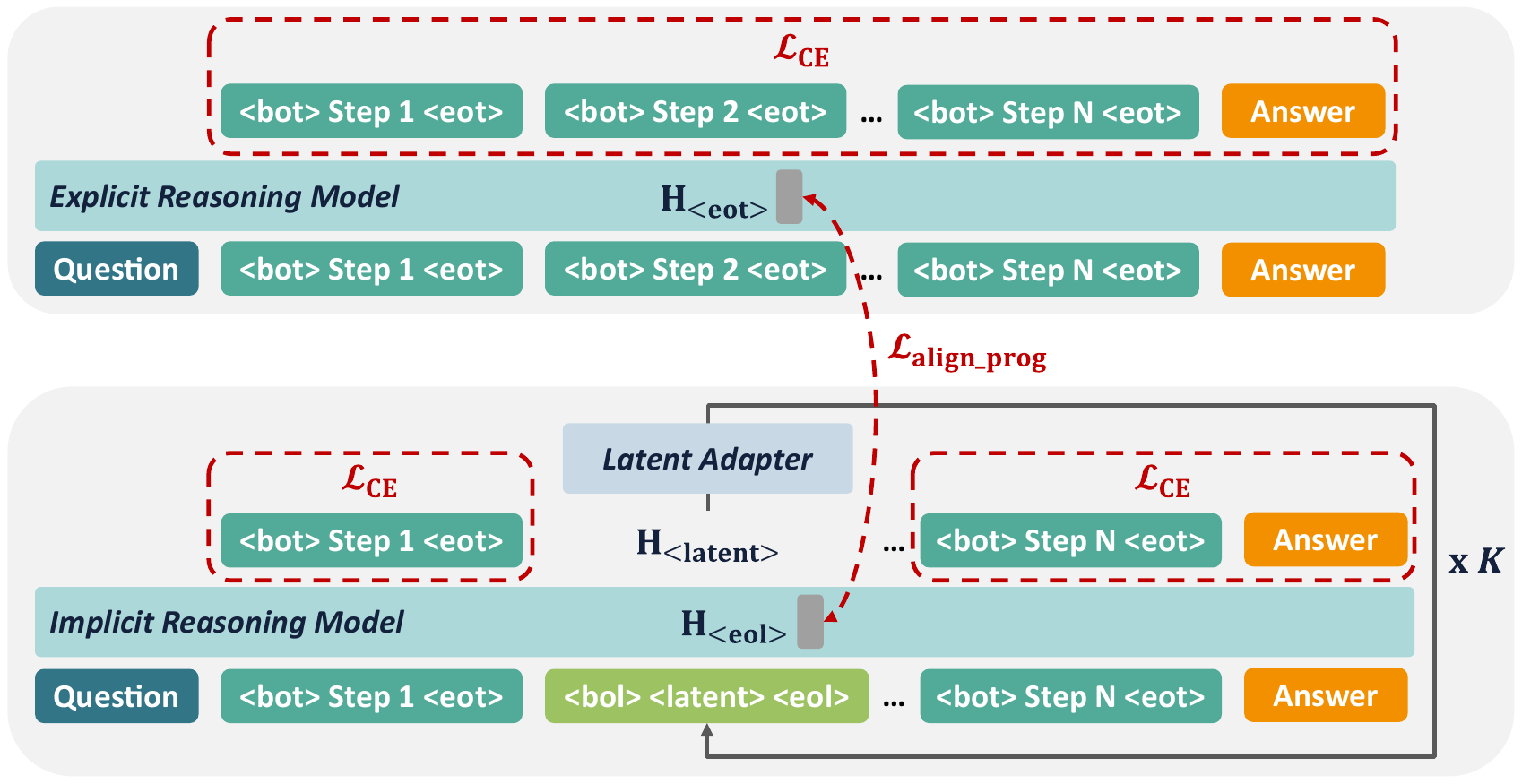}
  \caption{Training process of SpiralThinker. \texttt{Step} indicates a textual step, and \texttt{<latent>} indicates a latent step. Only one \texttt{<latent>} token is illustrated for clarity.}
  \label{fig:specific framework}
\end{figure*}

Extending this idea to the logit level, researchers have observed that a model’s output distributions already encode partial reasoning information \cite{wang2024chain}. This insight motivated approaches that either reuse logits directly as next-token proxies \cite{hao2024training} or use their probability weights to blend token embeddings \cite{zhang2025soft}, both aiming to establish a continuous information flow by feeding output logits back as input.

However, previous methods either treat latent representations as isolated tokens processed in a single forward pass or compress explicit reasoning into static approximations. In contrast, our approach models multiple latent representations as an integrated reasoning step and applies iterative computation to achieve deeper and more coherent latent reasoning.

\subsection{Iterative Reasoning}
Iterative computation, the idea of refining representations through repeated updates, is a long-standing concept in machine learning and has been widely employed in architectures such as recurrent neural networks \cite{goodfellow2016deep}. Building on the Universal Transformer \cite{dehghani2018universal}, which demonstrated that incorporating recurrence into the Transformer architecture \cite{vaswani2017attention} enhances performance on both algorithmic and language understanding tasks, subsequent research has further shown that iterative computation facilitates implicit reasoning. 

Recent studies have explored various ways to integrate recurrence while preserving the Transformer architecture, emphasizing iterative processing with text tokens rather than latent representations. Some approaches treat the entire model as an iterative unit that refines hidden states across multiple iterations, thereby enabling implicit reasoning within hidden states \cite{saunshi2025reasoning}. Other methods introduce block-level recurrence, reusing the same block during both training and inference to balance computational cost and reasoning accuracy \cite{geiping2025scaling, yu2025enhancing}. A complementary line of work applies recurrence selectively within layers, assigning additional iterative updates to difficult tokens without increasing the overall parameter count \cite{chen-etal-2025-inner}.

Overall, recurrence represents more than additional computation, as it naturally aligns with stepwise reasoning. Theoretically, an iterative model with $T$ iterations can simulate a $T$-step reasoning process, establishing a one-to-one correspondence between the iterative process and the reasoning steps \cite{saunshi2025reasoning}.

However, existing iterative approaches typically rely solely on the standard language modeling objective, which offers no direct supervision over the latent reasoning dynamics. In contrast, our work introduces a progressive alignment objective that explicitly guides and stabilizes latent reasoning throughout the iterative process.

\section{Methodology}
This section presents the overall framework of SpiralThinker (Figure~\ref{fig:specific framework}). We begin with the explicit reasoning stage (§\ref{subsec:explicit_reasoning}), which establishes stepwise reasoning capability, and then detail the implicit reasoning stage, consisting of the iterative process (§\ref{subsubsec:iterative_process}) and the latent adapter (§\ref{subsubsec:latent_adapter}). Next, we describe the alignment and total training objectives (§\ref{subsubsec:alignment_objective}), followed by the inference strategy (§\ref{subsec:inference_strategy}), which explains how SpiralThinker alternates between explicit and implicit reasoning during generation.

\subsection{Explicit Reasoning}
\label{subsec:explicit_reasoning}
We begin by fine-tuning the model under an explicit reasoning setting, which equips it with fundamental step-by-step reasoning capability and lays the foundation for the subsequent implicit reasoning stage.

Formally, given a question $Q$, the target sequence $R$ comprises two components: (i) a sequence of reasoning steps and (ii) the final answer. During training, the model learns to generate both elements conditioned on $Q$. The training objective is defined as:
\begin{equation}
\label{eq:LM_loss}
    \mathcal{L}_{\mathrm{CE}} = -\log P(R \mid Q)
\end{equation}

Each reasoning step is enclosed by boundary tokens \texttt{<bot>} (begin-of-text) and \texttt{<eot>} (end-of-text), which later facilitate alignment with latent representations.

\subsection{Implicit Reasoning}
\subsubsection{Data Construction}
\label{subsubsec:data_construction}
To enable the model to transition between textual and latent reasoning within a unified framework, we construct an alternating text–latent scheme. Specifically, every textual reasoning step at odd or even positions is replaced by $N$ latent tokens \texttt{<latent>}, forming a latent step while keeping the final answer unchanged. This scheme allows textual and latent reasoning to coexist within the same sequence, enabling the model to learn transitions between the two modes during training. 

As with textual steps, each latent step is delimited by \texttt{<bol>} (begin-of-latent) and \texttt{<eol>} (end-of-latent). The complete data format is described in Appendix~\ref{sec:data_format}.

\subsubsection{Iterative Process}
\label{subsubsec:iterative_process}
Given the constructed scheme, we denote the input as $\mathbf{x} = [x_1, \ldots, x_T]$, where each token $x_t$ represents either a textual or latent token. Each token is projected into the embedding space through an embedding matrix $E \in \mathbb{R}^{V \times d}$:
\begin{equation}
    \mathbf{e}_t = E[x_t]
\end{equation}
where $V$ denotes the vocabulary size and $d$ the hidden dimension. The resulting embedding sequence $\mathbf{E} = [\mathbf{e}_1, \ldots, \mathbf{e}_T]$ is then passed through a stack of $L$ layers:
\begin{equation}
    \mathbf{H}^{(l)} = f^{(l)}(\mathbf{H}^{(l-1)}), 
    \qquad \mathbf{H}^{(0)} = \mathbf{E}
\end{equation}
yielding final hidden states $\mathbf{H}^{(L)} = [\mathbf{h}_1^{(L)}, \ldots, \mathbf{h}_T^{(L)}]$.

We then perform iterative updates over latent representations. At iteration $k$ ($k=1,\dots,K$), we extract the hidden states corresponding to latent tokens, denoted as $\mathbf{H}^{(L,k-1)}_{\text{\texttt{<latent>}}}$, and transform them using a mapping module $g_{\phi}(\cdot)$:
\begin{equation}
    \tilde{\mathbf{H}}^{(k)}_{\text{\texttt{<latent>}}} = g_{\phi}\!\left(\mathbf{H}^{(L,k-1)}_{\text{\texttt{<latent>}}}\right)
\end{equation}
The transformed states are then written back into their corresponding positions in the embedding sequence to form an updated input $\mathbf{E}^{(k)}$. A subsequent forward pass conditioned on $\mathbf{E}^{(k)}$ produces updated hidden states $\mathbf{H}^{(L,k)}$. Repeating this procedure for $K$ iterations enables continuous reasoning over latent representations and fosters coherent integration of information between latent and contextual representations.

\subsubsection{Latent Adapter}
\label{subsubsec:latent_adapter}
Since final-layer hidden states and input embeddings reside in distinct subspaces~\cite{zhang2025soft}, we implement $g_{\phi}(\cdot)$ as a lightweight adapter that aligns hidden states with the embedding scale. This adapter projects latent representations back into the embedding space before rewrite, ensuring compatibility between latent and textual representations.

Formally, the adapter comprises a residual multilayer perceptron followed by RMSNorm and scaling:
\begin{equation}
    \tilde{\mathbf{h}} = \mathrm{norm}\bigl(\mathbf{h} + W_{2}\,\mathrm{SiLU}(W_{1}\mathbf{h})\big) \cdot \mathrm{target\_rms}
\end{equation}
where $W_1, W_2 \in \mathbb{R}^{d \times d}$ are learnable parameters, and $\mathrm{SiLU}(\cdot)$ denotes the SiLU activation function.

The fixed scalar factor $\mathrm{target\_rms}$ is derived from the pretrained embedding matrix, excluding the newly added special tokens, as
\begin{equation}
    \mathrm{target\_rms} = \mathbb{E}_{i \in V_{\text{base}}}\left[\sqrt{\frac{1}{d}\sum_{j=1}^{d} E_{ij}^{2}}\right]
\end{equation}
where $V_{\text{base}}$ denotes the pretrained vocabulary size. This ensures that the mapped latent representations remain consistent with the base embedding distribution.

\subsubsection{Alignment Objective}
\label{subsubsec:alignment_objective}
Although the iterative process enables repeated updates of latent representations, it does not inherently guarantee that these representations encode meaningful reasoning trajectories. We therefore introduce an alignment objective that constrains each latent representation to align with its corresponding textual counterpart.

\paragraph{Intra-iteration alignment}
For each training instance, we perform a forward pass to obtain hidden states at \texttt{<eol>}, and use a frozen model trained under the explicit reasoning setting (§\ref{subsec:explicit_reasoning}) to extract the corresponding \texttt{<eot>} hidden states for each textual reasoning step. 
Since the implicit model is initialized from the same explicitly trained backbone, latent and textual representations start from a shared reasoning-aware representation space, which helps reduce distributional mismatch during implicit reasoning training.
These tokens occur immediately after latent and textual reasoning steps, respectively, and serve as compact summaries of their preceding reasoning content.
Importantly, the aligned hidden states are not isolated step representations: they are prefix-conditioned states of the underlying language model, encoding cumulative information from all preceding reasoning steps.

Formally, let $\mathbf{H}^{(l)}_{\texttt{<eol>}}$ and $\mathbf{H}^{(l)}_{\texttt{<eot>}}$ denote the layer-$l$ hidden states at \texttt{<eol>} and \texttt{<eot>}, respectively. The alignment objective is defined as
\begin{equation}
\label{eq:align_base}
    \mathcal{L}_{\mathrm{align}} = \frac{1}{L}
    \sum_{l=1}^{L}
    \frac{\big\lVert \mathbf{H}^{(l)}_{\texttt{<eol>}} - \mathbf{H}^{(l)}_{\texttt{<eot>}} \big\rVert_{1}}{\sigma^{(l)}}
\end{equation}
where $\sigma^{(l)}$ is a per-layer normalization factor estimated from the frozen model as the average feature-wise standard deviation of hidden states across textual steps. This normalization maintains scale consistency across layers and stabilizes optimization~\cite{shen2025codi}. 

By aligning hidden states at these positions, the objective provides indirect supervision for latent tokens. Although these tokens are not directly optimized, the alignment loss propagates informative gradients through their hidden states, encouraging them to encode reasoning information consistent with explicit reasoning steps while operating over a continuously evolving reasoning state rather than enforcing strict equivalence on isolated steps.

\paragraph{Progressive alignment}
While the intra-iteration alignment objective~\eqref{eq:align_base} aligns representations within a single pass, the iterative process introduces an additional temporal dimension across multiple iterations. During this process, earlier iterations are expected to explore diverse reasoning trajectories, whereas later iterations should progressively consolidate these into accurate reasoning steps. To capture this progression, we compute the alignment loss at each iteration and aggregate them using a weighting schedule that emphasizes later iterations.

Specifically, we define a softmax-based weighting vector that increases monotonically with the iteration index:
\begin{equation}
\label{eq:vr}
    \mathbf{v} = \mathrm{softmax}\big(\alpha [1,\ldots,K]\big), \qquad \alpha>0
\end{equation}
where $\alpha$ is adaptively scaled with the number of iterations, assigning greater weights to later iterations. 
This weighting schedule increases the contribution of later iterations, encouraging the model to prioritize more consolidated reasoning states. At the same time, it preserves signals from earlier exploratory passes, thereby providing trajectory-level guidance rather than enforcing strict step-wise equivalence across iterations.
The progressive alignment objective is therefore defined as
\begin{equation} 
\label{eq:align_progressive} 
    \mathcal{L}_{\mathrm{align\_prog}} = \sum_{k=1}^{K} v_k\,\mathcal{L}_{\mathrm{align}}^{(k)} 
\end{equation}

\begin{table*}[t]
\renewcommand{\arraystretch}{1}
    \centering
    \begin{tabular}{lccc}
        \Xhline{1.0pt}
        \textbf{Methods} & \textbf{GSM8K-Aug (\%)} & \textbf{ProsQA (\%)} & \textbf{StrategyQA (\%)}\\ \Xhline{0.5pt}
        iCoT-KD \citep{deng2023implicit}        & 24.11             & 98.00             & \underline{62.88}\\ 
        iCoT-SI \citep{deng2024explicit}        & 29.72             & \underline{99.00} & 60.69\\ 
        Token Assorted \citep{su2025token}$^a$    & 48.70             & 96.20             & -\\
        Coconut \citep{hao2024training}         & 49.85             & 97.80             & 60.00\\
        CODI \citep{shen2025codi}               & 51.02             & 80.80             & 60.70\\
        Pause Token \citep{goyal2024think}      & \underline{53.37} & 95.80             & 57.64\\
        \cdashline{1-4}
        \textbf{SpiralThinker}                  & \textbf{56.56}    & \textbf{99.40}    & \textbf{63.32}\\ 
        \Xhline{1.0pt}
        \multicolumn{4}{l}{\footnotesize{$^a$ As the official code is not publicly released, the baseline results are taken from the papers.}}
    \end{tabular}
    \caption{Accuracy (\%) of SpiralThinker compared with baselines. \textbf{Bold} numbers indicate the best performance, and \underline{underlined} numbers indicate the second-best.}
    \label{tab:overall-performance}
\end{table*}

\paragraph{Total objective}
Integrating all components described above, we formulate the final training objective as 
\begin{equation}
\label{eq:total_objective}
    \mathcal{L}_{\mathrm{total}} = \mathcal{L}_{\mathrm{CE}} + \lambda\,\mathcal{L}_{\mathrm{align\_prog}}
\end{equation}
where $\lambda$ controls the weight of the progressive alignment term. Because the \texttt{<latent>} token has no explicit surface form, its prediction is excluded from the supervision signal in $\mathcal{L}_{\mathrm{CE}}$.

\begin{table*}[t]
\renewcommand{\arraystretch}{1.0}
    \centering
    \begin{tabular}{ccccc}
        \Xhline{1.0pt}
        \textbf{Alignment} & \textbf{Iteration} & \textbf{GSM8K-Aug (\%)} & \textbf{ProsQA (\%)} & \textbf{StrategyQA (\%)}\\ \Xhline{0.5pt}
        \textcolor{red}{\ding{55}}       & \textcolor{red}{\ding{55}}       & 45.49                 & 98.00             & 59.39\\ 
        \textcolor{darkgreen}{\ding{51}} & \textcolor{red}{\ding{55}}       & 48.67 \small{(+3.18)} & 98.60 \small{(+0.60)} & 61.14 \small{(+1.75)}\\ 
        \textcolor{red}{\ding{55}}       & \textcolor{darkgreen}{\ding{51}} & 45.72 \small{(+0.23)} & 97.40 \small{(-0.60)} & 58.08 \small{(-1.31)}\\ 
        \textcolor{darkgreen}{\ding{51}} & \textcolor{darkgreen}{\ding{51}} & \textbf{56.56 \small{(+11.07)}} & \textbf{99.40 \small{(+1.40)}} & \textbf{63.32 \small{(+3.93)}}\\ 
        \Xhline{1.0pt}
    \end{tabular}
    \caption{Ablation study on the effect of applying the iterative process and alignment objective.}
    \label{tab:ablation}
\end{table*}

\subsection{Inference Strategy}
\label{subsec:inference_strategy}
During inference, we prepend $N$ \texttt{<latent>} tokens to each input question. The model first performs implicit reasoning over these latent tokens, and subsequently produces explicit textual reasoning steps conditioned on the updated latent representations. Owing to the alternating training scheme, it learns to autonomously insert $N$ \texttt{<latent>} tokens after each textual step, thereby alternating between implicit and explicit reasoning throughout the generation process. This interleaving continues until the model generates the final answer or reaches the predefined generation limit. All results are obtained using greedy decoding to ensure deterministic reasoning trajectories.

\section{Experimental Setup}
\subsection{Dataset \& Metrics}
We evaluate SpiralThinker on three reasoning benchmarks—GSM8K-Aug, ProsQA, and StrategyQA—which target mathematical, logical, and commonsense reasoning, respectively. These datasets collectively span diverse reasoning domains, requiring both reasoning capability and relevant knowledge to achieve competitive performance. Detailed dataset descriptions are provided in Appendix~\ref{sec:datasets}.

Performance is measured using answer-level accuracy, defined as the proportion of instances whose final predicted answer exactly matches the ground-truth label across all benchmarks.

\subsection{Baselines}
We compare SpiralThinker with recent latent reasoning approaches, retraining all publicly available implementations under identical model architectures and datasets to ensure a fair comparison. Comprehensive descriptions of all baselines are given in Appendix~\ref{sec:baselines}.

\subsection{Implementation Details}
We adopt Llama-3.2-1B \citep{grattafiori2024llama} as the backbone and apply LoRA fine-tuning \cite{hu2022lora} to both SpiralThinker and all baselines. Training is conducted on four A100 GPUs using the Hugging Face library \citep{wolf2019huggingface}, with initialization from publicly released pretrained weights\footnote{https://huggingface.co/meta-llama/Llama-3.2-1B}. Complete hyperparameter configurations are provided in Appendix~\ref{sec:hyperparameter}.

\section{Results and Analysis}
\subsection{Overall Performance}
\label{subsec:overall_performance}
To evaluate the overall reasoning capability of SpiralThinker, we compare it with baselines on mathematical, logical, and commonsense reasoning benchmarks. As shown in Table~\ref{tab:overall-performance}, SpiralThinker consistently outperforms prior methods across all tasks. On GSM8K-Aug, it achieves 56.56\%, surpassing the previous best by 3.19 points, indicating that the iterative process particularly enhances multi-step numerical reasoning. On ProsQA, it reaches 99.40\%, surpassing the previous best by 0.40 points and suggesting better stability in high-accuracy regimes. On StrategyQA, the score of 63.32\% exceeds the previous best by 0.44 points, reflecting stronger commonsense reasoning.

Taken together, these cross-benchmark improvements suggest that modeling implicit reasoning as an iterative process over latent representations yields generalizable gains beyond specific tasks.

\subsection{Ablation Study}
\label{subsec:ablation_study}
To disentangle the contributions of SpiralThinker’s two core components—the iterative process and the progressive alignment objective—we conducted a series of ablations. We began with a single-iteration baseline employing latent tokens but trained only with the cross-entropy objective. We then added the alignment objective to encourage meaningful latent representations. Next, we incorporated the iterative process while retaining only the cross-entropy objective, isolating the effect of iteration. Finally, we combined both components to evaluate their joint effect.

As shown in Table~\ref{tab:ablation}, employing the iterative process alone yields limited improvement and can even degrade performance, suggesting that iterative updates in latent representation space may drift when left unconstrained. In contrast, incorporating the alignment objective—either in the single-pass or iterative setting—consistently enhances results, indicating that structured guidance is important for stabilizing latent reasoning. When both mechanisms are combined, the gains become substantial, confirming that iteration and progressive alignment play complementary roles: iteration provides additional reasoning depth, while progressive alignment regulates this multi-step evolution to keep it structured, stable, and coherent.

\subsection{Analysis Study}
\label{subsec:analysis_study}
We further examine the influence of three key factors: the number of latent tokens, the number of iterations, and the weighting schedule of the progressive alignment objective on the performance of SpiralThinker. This analysis provides deeper insights into the model’s sensitivity and internal dynamics.

\paragraph{The effect of number of latent tokens}
We first examined how varying the number of latent tokens ($N \in [1,10]$) influences overall performance. To control for confounding factors, this experiment excluded iteration while retaining the alignment objective.

\begin{figure}[t]
  \centering
  \includegraphics[width=1.0\linewidth]{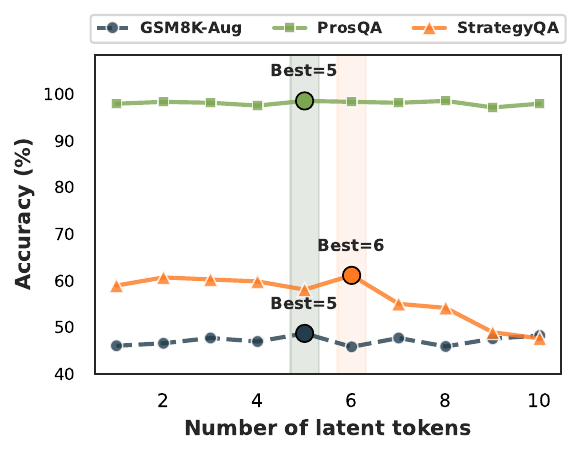}
  \caption{Accuracy on different datasets as the number of latent tokens varies.}
  \label{fig:number_latent}
\end{figure}

As shown in Figure~\ref{fig:number_latent}, performance on StrategyQA decreases when $N>6$, whereas accuracy on other datasets remains largely stable across the range $N \in [1,10]$. This trend is consistent with the observations reported by \citet{goyal2024think} and \citet{shen2025codi}, indicating that the optimal number of latent tokens depends on the dataset and reflects inherent differences in reasoning complexity. 

Overall, the best results are obtained with $N{=}5$ on GSM8K-Aug and ProsQA, and with $N{=}6$ on StrategyQA; these configurations are used in other experiments.

\begin{figure*}[t]
  \centering
  \includegraphics[width=1.0\linewidth]{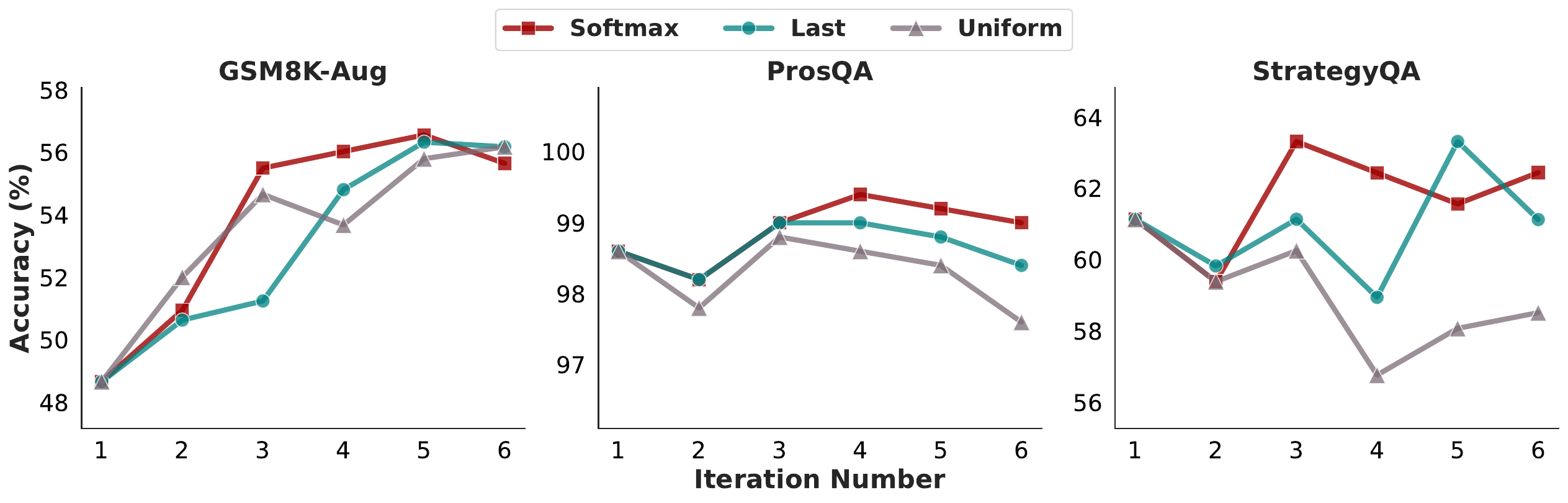}
  \caption{Accuracy on different datasets as the number of iterations varies.}
  \label{fig:number_iteration}
\end{figure*}

\begin{figure*}[t]
  \centering
  \includegraphics[width=1.0\linewidth]{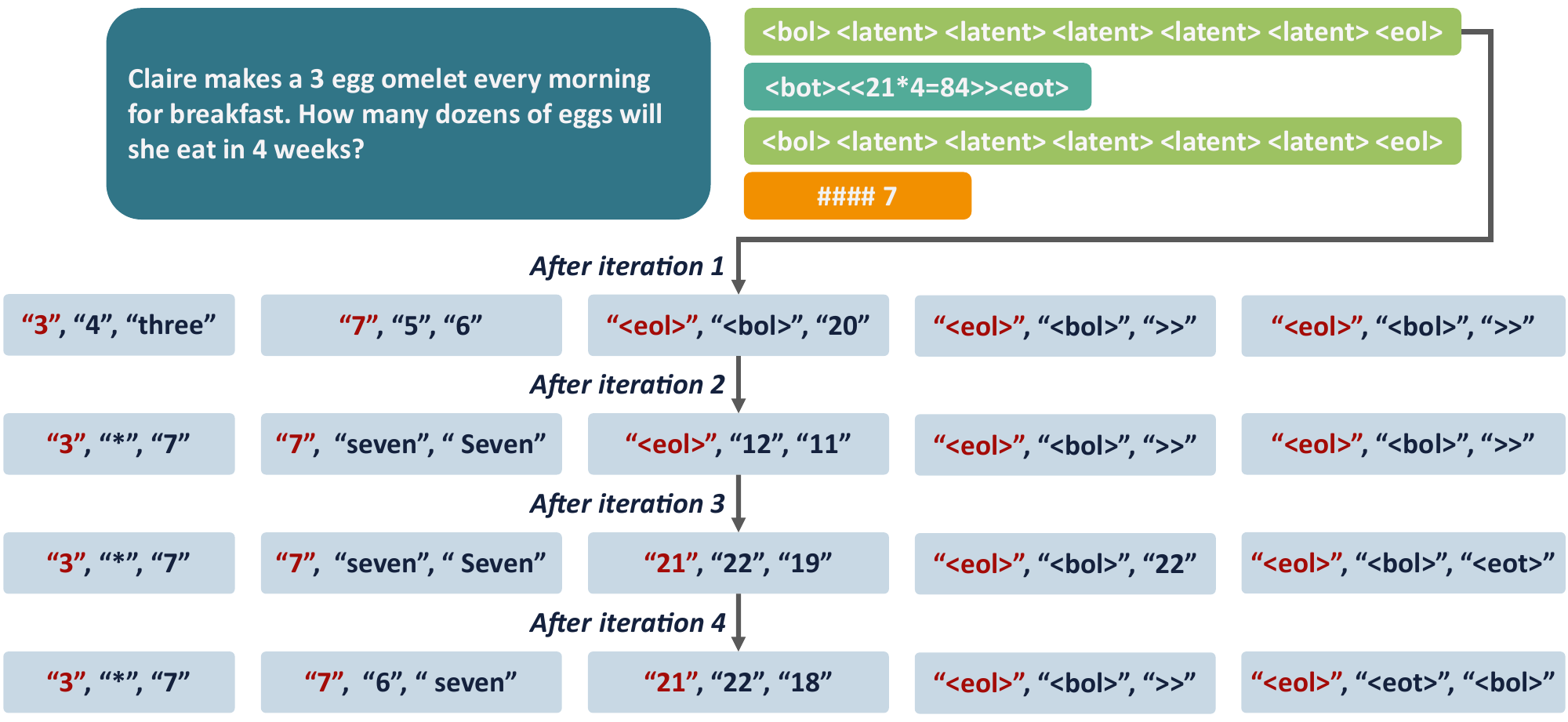}
  \caption{The upper part shows the reasoning steps generated by SpiralThinker for a sample problem, while the lower part presents the top three tokens most similar to each latent representation at the first latent step during the iterative process. The top-ranked token is highlighted in \textcolor{darkred}{\textbf{red}}.}
  \label{fig:qualitative_analysis}
\end{figure*}

\paragraph{The effect of number of iterations}
Next, we analyze how the number of iterations ($K \in [1,6]$) in the iterative process affects performance.

As illustrated in Figure~\ref{fig:number_iteration}, a moderate number of iterations consistently improves accuracy, while excessive iteration leads to saturation or slight degradation—most notably on GSM8K-Aug and ProsQA. Although StrategyQA exhibits minor fluctuations, it follows a similar trend, suggesting that increasing iterations does not necessarily enhance reasoning performance.

Accordingly, when using the softmax-based weighting schedule, the best results are achieved with $K{=}5$ on GSM8K-Aug, $K{=}4$ on ProsQA, and $K{=}3$ on StrategyQA, again suggesting that the optimal number of iterations is dataset-dependent. These settings are adopted in other experiments.

\paragraph{The effect of the weighting schedule}
Finally, we investigated the effect of different weighting schedules for the progressive alignment objective. Specifically, we compared (1) uniform weighting across iterations, (2) weighting only the last iteration, and (3) our proposed softmax-based scheduler.

As illustrated in Figure~\ref{fig:number_iteration}, uniform weighting consistently underperforms, with the effect most pronounced on ProsQA and StrategyQA. In contrast, both last-iteration-only weighting and the softmax-based weighting schedule improve performance, with the latter generally performing best: it achieves the highest peak accuracy on GSM8K-Aug and ProsQA, and converges to its performance plateau with fewer iterations on StrategyQA. These findings highlight that maintaining appropriately weighted alignment across iterations is crucial for effective latent reasoning during the iterative process.

\subsection{Qualitative Analysis of \texttt{<latent>}}
To assess how SpiralThinker utilizes latent representations during the iterative process, we conduct a qualitative analysis of the \texttt{<latent>} tokens. Since the latent adapter aligns latent representations with token embeddings, we identify tokens whose embeddings are most similar to these latent representations. Using a GSM8K-Aug test instance as a case study, we report the top three tokens associated with each latent representation across iterations.

As shown in Figure~\ref{fig:qualitative_analysis}, the first and second latent tokens consistently capture relevant reasoning factors throughout iterations. Notably, the first latent token encodes the multiplication operator “*”, indicating sensitivity to arithmetic operators. The third latent token primarily stores intermediate results, which progressively converge to the correct values. The fourth and fifth latent tokens frequently produce the \texttt{<eol>} symbol, marking reasoning termination. Collectively, these observations indicate that the latent representations appear to capture reasoning-relevant signals that evolve coherently throughout the iterative process.

\section{Conclusion}
We presented SpiralThinker, a framework that bridges iterative computation and latent reasoning through repeated updates over latent representations. By incorporating a progressive alignment objective and structured annotations, SpiralThinker maintains consistency between latent and textual reasoning, enabling extended implicit reasoning without generating additional tokens. Experiments on GSM8K-Aug, ProsQA, and StrategyQA demonstrate that SpiralThinker consistently surpasses previous latent reasoning approaches. Ablation results further confirm that both iteration and alignment are essential for stable and coherent reasoning, and appropriate alignment proves critical for an effective iterative process. In conclusion, SpiralThinker bridges iterative computation and latent reasoning, demonstrating that aligned iterative updates can reliably steer latent reasoning.

\section*{Limitations}
While SpiralThinker achieves competitive performance, there remain two main areas for improvement.

First, SpiralThinker currently applies a fixed number of iterations across all reasoning steps, regardless of their difficulty. A more desirable approach would dynamically adjust the number of iterations based on instance difficulty—using fewer iterations for easy steps and more for challenging ones. In practice, this sensitivity could potentially be mitigated through lightweight calibration on a small held-out set or adaptive stopping criteria based on convergence or confidence signals.

Second, although the proposed design enables interleaved reasoning between textual and latent representations through its data scheme and training strategy, such interleaving may not be necessary at every reasoning step. Future work could investigate adaptive mechanisms that allow the model to decide when latent reasoning should be activated.

\section*{Acknowledgments}
This work was supported by the National Research Foundation (NRF) funded by the Korean Government (MSIT) (No. RS-2025-00523472); by an IITP grant funded by the Korean Government (MSIT) (No. RS-2020-II201361, Artificial Intelligence Graduate School Program (Yonsei University)); and by the Regional Innovation System \& Education (RISE) program through the Jeju RISE Center, funded by the Ministry of Education (MOE) and the Jeju Special Self-Governing Province, Republic of Korea (2025-RISE-17-003).

\bibliography{custom}

\appendix

\section{Data Format}
\label{sec:data_format}

\begin{tcolorbox}[
    colback=gray!5!white,
    colframe=black,
    title=\textbf{\textsc{Data Format}},
    width=\linewidth, 
    boxrule=1pt
]
\small
\textbf{Explicit Reasoning:}
\begin{verbatim}
    Question
    <bot> Step 1 <eot>
    <bot> Step 2 <eot>
    <bot> Step 3 <eot>
    <bot> Step 4 <eot>
    #### Answer
\end{verbatim}
\textbf{Implicit Reasoning:}
\begin{verbatim}
    (1) Question
    <bol> N x <latent> <eol>
    <bot> Step 2 <eot>
    <bol> N x <latent> <eol>
    <bot> Step 4 <eot>
    #### Answer
    
    (2) Question
    <bot> Step 1 <eot>
    <bol> N x <latent> <eol>
    <bot> Step 3 <eot>
    <bol> N x <latent> <eol>
    #### Answer
\end{verbatim}
\end{tcolorbox}
\begin{minipage}{0.45\textwidth}
\captionof{table}{Data formats used for explicit and implicit reasoning.}
\label{tab:data_format}
\end{minipage}
\noindent Table~\ref{tab:data_format} summarizes the data formats used for explicit and implicit reasoning.

\section{Datasets}
\label{sec:datasets}
\paragraph{GSM8K-Aug} GSM8K \cite{cobbe2021training} is a benchmark of grade-school math problems widely used to evaluate multi-step numerical reasoning involving both language understanding and mathematical computation. GSM8K-Aug \citep{deng2023implicit} is an augmented version generated with GPT-4 based on the original GSM8K training set.

\paragraph{ProsQA} ProsQA \citep{hao2024training} is a synthetic question–answering dataset designed to evaluate logical reasoning capability. It is constructed from randomly generated directed acyclic graphs that specify the known conditions and reasoning dependencies. Each instance requires multi-step planning to identify the correct reasoning chain, thereby assessing the model's capability to perform structured logical reasoning.

\paragraph{StrategyQA} StrategyQA \citep{geva-etal-2021-aristotle} is a binary (yes/no) question-answering dataset used to evaluate commonsense reasoning across diverse topics. It challenges models to perform multi-step reasoning by decomposing complex questions into intermediate steps before producing the final answer.

Since the original dataset lacks textual reasoning steps, we use Llama-3.2-7B \citep{grattafiori2024llama} with a few-shot prompting setup (Table~\ref{tab:prompt}) following chain-of-thought \citep{wei2022chain} to generate them. Moreover, because the official test set is unavailable, we follow prior work by using the validation set for testing and sampling an equal-sized subset from the training data as the new validation set.

The dataset statistics are summarized in Table~\ref{tab:dataset_statistic}.

\begin{table*}[t]
    \centering
    \begin{tabular}{lcccc}
        \Xhline{1.0pt}
        \textbf{Dataset} & \textbf{\# Train (explicit)} & \textbf{\# Train (implicit)}& \textbf{\# Validation} & \textbf{\# Test} \\ \Xhline{0.5pt}
        GSM8K-Aug  & 384,620 & 643,762 & 500 & 1,319 \\
        ProsQA  & 17,886 & 35,772 & 300 & 500 \\
        StrategyQA & 1,714 & 3,412 & 229 & 229 \\
        \Xhline{1.0pt}
    \end{tabular}
    \caption{Dataset statistics.}
    \label{tab:dataset_statistic}
\end{table*}

\begin{table*}[t]
    \centering
    \begin{tabular}{lccc}
        \Xhline{1.0pt}
        \textbf{Models} & \textbf{GSM8K-Aug (\%)} & \textbf{ProsQA (\%)} & \textbf{StrategyQA (\%)}\\ \Xhline{0.5pt}
        Llama-3.2-1B                  & 56.56 & 99.40 & 63.32 \\
        Qwen2.5-0.5B                  & 46.85 & 96.80 & 58.95 \\ 
        Qwen2.5-1.5B                  & 58.45 & 99.40 & 65.07 \\ 
        \Xhline{1.0pt}
    \end{tabular}
    \caption{Accuracy (\%) of SpiralThinker across backbone families and model sizes.}
    \label{tab:method_scalability}
\end{table*}

\section{Baselines}
\label{sec:baselines}
\paragraph{Pause Token} \citep{goyal2024think}
This method introduces pause training, which allows a language model to perform additional computation before generating the reasoning process. During both training and inference, special \texttt{<pause>} tokens are inserted to delay token generation, enabling the model to refine its hidden representations prior to producing the next token.

\paragraph{ICoT-KD} \citep{deng2023implicit}
This approach proposes implicit chain-of-thought reasoning, in which reasoning occurs within hidden states rather than through explicit text. It comprises three stages: (1) a student model learns from a teacher’s hidden states during intermediate reasoning; (2) an emulator predicts the teacher’s hidden states directly from inputs through layer-wise reasoning; and (3) the emulator and student are jointly trained to produce final answers without explicit reasoning steps.

\paragraph{ICoT-SI} \citep{deng2024explicit}
This method introduces a curriculum-based framework for transforming explicit reasoning into implicit reasoning. Starting with a model trained via explicit chain-of-thought supervision, it gradually removes intermediate steps during fine-tuning, encouraging the model to internalize reasoning within its hidden states.

\begin{table*}[t]
    \centering
    \small
    \renewcommand{\arraystretch}{1.1}
    \begin{tabular}{llcccccc}
        \Xhline{1.0pt}
        \textbf{Benchmark} & \textbf{Method} & \textbf{Acc.} & \makecell{\textbf{Avg. textual}\\\textbf{tokens}} & \makecell{\textbf{Text}\\\textbf{reduction}} & \makecell{\textbf{Avg. latent}\\\textbf{steps}} & \makecell{\textbf{Avg. latent}\\\textbf{iters}} & \makecell{\textbf{Total forward}\\\textbf{count}} \\
        \Xhline{0.4pt}
        \multirow{2}{*}{GSM8K-Aug} 
        & Explicit      & 62.02\% & 30.56 & --      & 0.00 & 0 & 30.56 \\
        & SpiralThinker & 56.56\% & 13.99 & 54.22\% & 1.71 & 5 & 22.54 \\
        \Xhline{0.4pt}
        \multirow{2}{*}{ProsQA} 
        & Explicit      & 97.60\% & 44.71 & --      & 0.00 & 0 & 44.71 \\
        & SpiralThinker & 99.40\% & 33.57 & 24.92\% & 2.83 & 4 & 44.89 \\
        \Xhline{0.4pt}
        \multirow{2}{*}{StrategyQA} 
        & Explicit      & 64.19\% & 77.14 & --      & 0.00 & 0 & 77.14 \\
        & SpiralThinker & 63.32\% & 39.77 & 48.44\% & 2.13 & 3 & 46.16 \\
        \Xhline{1.0pt}
    \end{tabular}
    \caption{Comparison between explicit reasoning and SpiralThinker on computation cost.}
    \label{tab:computation_cost}
\end{table*}

\paragraph{Coconut} \citep{hao2024training}
This approach enables reasoning in a continuous latent space rather than through discrete text tokens. Instead of decoding output logits into tokens, the model iteratively feeds the final hidden state back as the next input embedding. Training follows a multi-stage curriculum that progressively replaces explicit reasoning steps with continuous latent representations.

\paragraph{Token Assorted} \citep{su2025token}
This hybrid framework integrates latent and textual reasoning tokens to enhance efficiency. A VQ-VAE encoder–decoder compresses the initial portion of a reasoning step into discrete latent tokens while retaining later steps as text, thereby reducing sequence length and computational cost.

\paragraph{CODI} \citep{shen2025codi}
This framework compresses explicit reasoning into a compact continuous latent representation. CODI jointly trains a teacher model with textual reasoning process and a student model using latent tokens, aligning their hidden states via a layer-wise loss. This self-distillation mechanism effectively transfers reasoning capability from the language space to the continuous latent space.

\section{Additional Experiments}
\label{sec:additional_experiments}
\subsection{Method Scalability}
We assess the scalability of SpiralThinker along two dimensions: backbone family and model size. To examine cross-family transferability, we replace the original Llama backbone with Qwen2.5, another decoder-only Transformer family. To study scaling with respect to model size, we further evaluate SpiralThinker on two Qwen2.5 variants, namely Qwen2.5-0.5B and Qwen2.5-1.5B.

As shown in Table~\ref{tab:method_scalability}, SpiralThinker can be effectively trained on both the Llama-3.2 and Qwen2.5 families, suggesting that the method is not restricted to a specific backbone and can extend to different decoder-only Transformer models. Moreover, within the Qwen2.5 family, performance consistently improves as the model size increases from 0.5B to 1.5B, yielding gains of 11.60 points on GSM8K-Aug, 2.60 points on ProsQA, and 6.12 points on StrategyQA. These results suggest that SpiralThinker exhibits favorable scalability across both model families and model sizes.

\subsection{Computational Cost Analysis}
To better understand the computational implications of SpiralThinker, we compare it with the explicit reasoning model (§\ref{subsec:explicit_reasoning}) on all benchmarks using the same Llama backbone. SpiralThinker is not designed as a reasoning acceleration method. Although latent reasoning can reduce the need to externalize reasoning as textual tokens, each latent reasoning step incurs additional forward passes. Consequently, lower visible verbosity does not necessarily imply lower inference cost.

Table~\ref{tab:computation_cost} reports both task performance and computation-related statistics. \textbf{Acc.} denotes task accuracy. \textbf{Avg. textual tokens} is the average number of textual tokens generated during inference. \textbf{Text reduction} measures the relative reduction in generated textual tokens for SpiralThinker compared with the explicit reasoning baseline. \textbf{Avg. latent steps} denotes the average number of latent reasoning steps generated by SpiralThinker. \textbf{Avg. latent iters} denotes the predefined number of iterative refinements applied to each latent reasoning step for a given benchmark.

To reflect the effective inference cost, we report the \textbf{Total forward count}. For the explicit reasoning model, generating one token corresponds to one forward pass; therefore, the total forward count is equal to the number of generated textual tokens. For SpiralThinker, the total forward count includes both the forward passes used to generate textual tokens and the additional forward passes introduced by the iterative process:
\begin{equation}
F_{\text{total}} = T_{\text{text}} + S_{\text{latent}} \cdot I_{\text{latent}} .
\end{equation}
Here, $F_{\text{total}}$ denotes the total forward count, $T_{\text{text}}$ the average number of generated textual tokens, $S_{\text{latent}}$ the average number of latent steps, and $I_{\text{latent}}$ the average number of latent iterations.

As shown in Table~\ref{tab:computation_cost}, SpiralThinker substantially reduces the number of generated textual tokens on all three benchmarks, with reductions of 54.22\% on GSM8K-Aug, 24.92\% on ProsQA, and 48.44\% on StrategyQA. However, its effect on total computation varies across datasets because latent reasoning introduces additional forward passes. On GSM8K, SpiralThinker reduces the total forward count from 30.56 to 22.54, indicating that the reduction in textual generation outweighs the cost of latent refinement. A similar trend is observed on StrategyQA, where the total forward count decreases from 77.14 to 46.16. In contrast, on ProsQA, although SpiralThinker generates fewer textual tokens, the additional cost of latent reasoning results in a nearly identical total forward count (44.89 vs.\ 44.71).

These results suggest that SpiralThinker should be viewed as a method that offers a controllable trade-off between computation and reasoning quality, rather than primarily optimized for efficiency. Its main advantage lies in reducing textual verbosity while preserving competitive reasoning performance, whereas the actual computational cost depends on the balance between fewer generated text tokens and the extra forward passes introduced by latent reasoning.

\section{Hyperparameter Settings}
\label{sec:hyperparameter}
The hyperparameters used in this study are summarized in Table~\ref{tab:hyperparameter}. Because the StrategyQA dataset is relatively small compared with the other datasets, slightly different hyperparameter settings were applied.

\section{Generation Results}
\label{sec:generation_results}
We report results on the GSM8K-Aug, ProsQA, and StrategyQA datasets, with two representative examples for each shown in Tables~\ref{tab:result_gsm8k}, \ref{tab:result_prosqa}, and \ref{tab:result_strategyqa}, respectively, to qualitatively illustrate SpiralThinker’s reasoning behavior across different domains.

\onecolumn

\begin{table}
    \centering
    \small
    \renewcommand{\arraystretch}{1.1}
    \begin{tabular}{ccl}
        \Xhline{1.0pt}
        \textbf{Hyperparameter} & \textbf{Value} & \textbf{Note}\\ \Xhline{0.5pt}
        \multicolumn{3}{c}{\textbf{\textit{Explicit Reasoning}}} \\ 
        epochs & 5/15 & Number of training epochs for GSM8K-Aug and ProsQA/StrategyQA. \\
        batch\_size & 128 & Batch size for GSM8K-Aug, ProsQA and StrategyQA. \\
        lr & $1\times10^{-4}$ & Learning rate for the AdamW optimizer. \\ \Xhline{0.5pt}
        \multicolumn{3}{c}{\textbf{\textit{Implicit Reasoning}}} \\
        epochs & 5/30 & Number of training epochs for GSM8K-Aug and ProsQA/StrategyQA. \\
        batch\_size & 128/32 & Batch size for GSM8K-Aug and ProsQA/StrategyQA. \\
        lr & $5\times10^{-5}$ & Learning rate for the AdamW optimizer. \\
         & 0.8 & Proportion of total weight assigned to the later iteration. \\
        $\lambda$ & 0.5 & Weight coefficient for the alignment objective term. \\ \Xhline{0.5pt}
        \multicolumn{3}{c}{\textbf{\textit{LoRA Configuration}}} \\
        $r$ & 32/16 & Rank for GSM8K-Aug and ProsQA/StrategyQA. \\
        $\mathrm{lora\_}\alpha$ & 64/32 & Scaling factor for GSM8K-Aug and ProsQA/StrategyQA. \\
        $\mathrm{lora\_dropout}$ & 0.05 & Dropout probability applied to LoRA layers. \\ 
        \Xhline{1.0pt}
    \end{tabular}
    \caption{Summary of hyperparameter settings used in the experiments.}
    \label{tab:hyperparameter}
\end{table}

\begin{tcolorbox}[
    colback=gray!5!white,
    colframe=black,
    title=\textbf{\textsc{Prompt for StrategyQA}},
    width=\textwidth,
    boxrule=1pt, 
    breakable
]
\textbf{1. Do hamsters provide food for any animals?}\newline
Hamsters are prey animals.\newline\
Prey are food for predators.\newline
Thus, hamsters provide food for some animals.\newline
So the answer is yes.
\newline\newline
\textbf{2. Could Brooke Shields succeed at University of Pennsylvania?}\newline
Brooke Shields went to Princeton University.\newline
Princeton University is about as academically rigorous as the University of Pennsylvania.\newline
Thus, Brooke Shields could also succeed at the University of Pennsylvania.\newline
So the answer is yes.
\newline\newline
\textbf{3. Hydrogen's atomic number squared exceeds number of Spice Girls?}\newline
Hydrogen has an atomic number of 1.\newline
1 squared is 1.\newline
There are 5 Spice Girls.\newline
Thus, Hydrogen’s atomic number squared is less than 5.\newline
So the answer is no.
\newline\newline
\textbf{4. Is it common to see frost during some college commencements?}\newline
College commencement ceremonies can happen in December, May, and June.\newline
December is in the winter, so there can be frost.\newline
Thus, there could be frost at some commencements.\newline
So the answer is yes.
\newline\newline
\textbf{5. Could a llama birth twice during War in Vietnam (1945-46)?}\newline
The War in Vietnam was 6 months.\newline
The gestation period for a llama is 11 months, which is more than 6 months.\newline
Thus, a llama could not give birth twice during the War in Vietnam.\newline
So the answer is no.
\newline\newline
\textbf{6. Would a pear sink in water?}\newline
The density of a pear is about $0.6g/cm^3$, which is less than water.\newline
Objects less dense than water float.\newline
Thus, a pear would float.\newline
So the answer is no.
\end{tcolorbox}
\begin{minipage}{\textwidth}
\captionof{table}{The prompt for StrategyQA dataset.}
\label{tab:prompt}
\end{minipage}

\begin{tcolorbox}[
    colback=gray!5!white,
    colframe=black,
    title=\textbf{\textsc{Result of GSM8K-Aug}},
    width=\textwidth,
    boxrule=1pt, 
    breakable
]
\textbf{1. Toulouse has twice as many sheep as Charleston. Charleston has 4 times as many sheep as Seattle. How many sheep do Toulouse, Charleston, and Seattle have together if Seattle has 20 sheep?}\newline
\textit{SpiralThinker:}
\begin{verbatim}
<bol><latent><latent><latent><latent><latent><eol>
<bot><<2*80=160>><eot>
<bol><latent><latent><latent><latent><latent><eol>
#### 260
\end{verbatim}
\textit{Ground Truth:}
\begin{verbatim}
<<20*4=80>>
<<2*80=160>>
<<20+160+80=260>>
#### 260
\end{verbatim}

\textbf{2. Claire makes a 3 egg omelet every morning for breakfast. How many dozens of eggs will she eat in 4 weeks?}\newline
\textit{SpiralThinker:}
\begin{verbatim}
<bol><latent><latent><latent><latent><latent><eol>
<bot><<21*4=84>><eot>
<bol><latent><latent><latent><latent><latent><eol>
#### 7
\end{verbatim}
\textit{Ground Truth:}
\begin{verbatim}
<<3*7=21>>
<<4*21=84>>
<<84/12=7>>
#### 7
\end{verbatim}
\end{tcolorbox}
\begin{minipage}{\textwidth}
\captionof{table}{Generated results of GSM8K-Aug.}
\label{tab:result_gsm8k}
\end{minipage}

\begin{tcolorbox}[
    colback=gray!5!white,
    colframe=black,
    title=\textbf{\textsc{Result of ProsQA}},
    width=\textwidth,
    boxrule=1pt, 
    breakable
]
\textbf{1. Every kerpus is a sterpus. Every vumpus is a gerpus. Rex is a impus. Rex is a vumpus. Every boompus is a terpus. Every shumpus is a zhorpus. Alex is a kerpus. Every terpus is a felpus. Bob is a zhorpus. Every fompus is a gerpus. Every yimpus is a jelpus. Every gwompus is a sterpus. Every gwompus is a zhorpus. Every yimpus is a kerpus. Alex is a gwompus. Every chorpus is a terpus. Every vumpus is a lempus. Every vumpus is a shumpus. Every shumpus is a boompus. Rex is a chorpus. Every impus is a fompus. Every chorpus is a impus. Every lempus is a boompus. Every vumpus is a fompus. Alex is a jelpus. Every jelpus is a scrompus. Every shumpus is a lempus. Every impus is a shumpus. Every chorpus is a zhorpus. Alex is a tumpus. Every gwompus is a yimpus. Alex is a yimpus. Is Rex a felpus or sterpus?}\newline
\textit{SpiralThinker:}
\begin{verbatim}
<bol><latent><latent><latent><latent><latent><eol>
<bot>Every chorpus is a terpus.<eot>
<bol><latent><latent><latent><latent><latent><eol>
#### Rex is a felpus.
\end{verbatim}
\textit{Ground Truth:}
\begin{verbatim}
Rex is a chorpus.
Every chorpus is a terpus.
Every terpus is a felpus.
#### Rex is a felpus.
\end{verbatim}
\textbf{2. Every impus is a lempus. Every timpus is a jompus. Every terpus is a shumpus. Every sterpus is a zhorpus. Every vumpus is a zhorpus. Eva is a timpus. Every rompus is a yumpus. Every yumpus is a lempus. Every yumpus is a impus. Every gorpus is a zhorpus. Every fompus is a quimpus. Rex is a vumpus. Every fompus is a zhorpus. Every zhorpus is a zumpus. Every sterpus is a yerpus. Every timpus is a impus. Every rompus is a timpus. Every lempus is a chorpus. Every vumpus is a fompus. Every zhorpus is a terpus. Every yerpus is a gwompus. Eva is a rompus. Every vumpus is a gorpus. Every vumpus is a sterpus. Every vumpus is a gwompus. Eva is a lempus. Is Rex a shumpus or jompus?}\newline
\textit{SpiralThinker:}
\begin{verbatim}
<bol><latent><latent><latent><latent><latent><eol>
<bot>Every vumpus is a zhorpus.<eot>
<bol><latent><latent><latent><latent><latent><eol>
<bot>Every terpus is a shumpus.<eot>
#### Rex is a shumpus.
\end{verbatim}
\textit{Ground Truth:}
\begin{verbatim}
Rex is a vumpus.
Every vumpus is a zhorpus.
Every zhorpus is a terpus.
Every terpus is a shumpus.
#### Rex is a shumpus.
\end{verbatim}
\end{tcolorbox}
\begin{minipage}{\textwidth}
\captionof{table}{Generated results of ProsQA.}
\label{tab:result_prosqa}
\end{minipage}

\begin{tcolorbox}[
    colback=gray!5!white,
    colframe=black,
    title=\textbf{\textsc{Result of StrategyQA}},
    width=\textwidth,
    boxrule=1pt, 
    breakable
]
\textbf{1. Is Miami a city on the American West Coast?}\newline
\textit{SpiralThinker:}
\begin{verbatim}
<bol><latent><latent><latent><latent><latent><latent><eol>
<bot>The American West Coast refers to the western coast of the United States, 
which includes states like California, Oregon, and Washington.<eot>
<bol><latent><latent><latent><latent><latent><latent><eol>
#### no
\end{verbatim}
\textit{Ground Truth:}
\begin{verbatim}
Miami is located in the state of Florida.
Florida is on the East Coast of the United States.
Thus, Miami is not on the West Coast.
#### no
\end{verbatim}
\textbf{2. Can amoebas get cancer?}\newline
\textit{SpiralThinker:}
\begin{verbatim}
<bol><latent><latent><latent><latent><latent><latent><eol>
<bot>Cancer is a disease that arises from abnormal cell growth and division.<eot>
<bol><latent><latent><latent><latent><latent><latent><eol>
<bot>Thus, amoebas cannot get cancer.<eot>
#### no
\end{verbatim}
\textit{Ground Truth:}
\begin{verbatim}
Cancer happens when cells in a multicellular organism grow uncontrollably and harm
the body.
Amoebas are single-celled organisms, so they don’t have tissues or organs that
cancer can affect.
Thus, amoebas cannot get cancer in the way animals or humans do.
#### no
\end{verbatim}
\end{tcolorbox}
\begin{minipage}{\textwidth}
\captionof{table}{Generated results of StrategyQA.}
\label{tab:result_strategyqa}
\end{minipage}

\end{document}